%% file: main.tex
\documentclass{article} % For LaTeX2e
\usepackage{iclr2019_conference,times}

\input{math_commands.tex}

\usepackage[utf8]{inputenc} % allow utf-8 input
\usepackage[T1]{fontenc}    % use 8-bit T1 fonts
\usepackage{hyperref}       % hyperlinks
\usepackage{url}            % simple URL typesetting
\usepackage{booktabs, makecell}       % professional-quality tables
\usepackage{amsfonts}       % blackboard math symbols
\usepackage{nicefrac}       % compact symbols for 1/2, etc.
\usepackage{microtype}      % microtypography
\usepackage{amsmath}
\usepackage{graphicx}
\usepackage{todonotes}
\usepackage{subcaption}
\usepackage{natbib}

\usepackage{algpseudocode,algorithmicx,algorithm}
  
\algrenewcommand\algorithmicrequire{\textbf{Precondition:}}  
\algrenewcommand\algorithmicensure{\textbf{Postcondition:}}

\usepackage{tabularx} % more advanced tables

\title{Reducing Adversarial Example \\ Transferability Using Gradient \\ Regularization}

\author{George Adam \\
Department of Computer Science \\
University of Toronto\\
\texttt{alex.adam@mail.utoronto.ca} 
\And
Petr Smirnov \\
Medical Biophysics \\
University of Toronto 
\AND
Benjamin Haibe-Kains \\
Medical Biophysics \\
University of Toronto
\AND
Anna Goldenberg \\
Department of Computer Science \\
University of Toronto
}

\begin{document}

\maketitle

\begin{abstract}
Deep learning algorithms have increasingly been shown to lack robustness to simple adversarial examples (AdvX). An equally troubling observation is that these adversarial examples transfer between different architectures trained on different datasets. We investigate the transferability of adversarial examples between models using the angle between the input-output Jacobians of different models. To demonstrate the relevance of this approach, we perform case studies that involve jointly training pairs of models. These case studies empirically justify the theoretical intuitions for why the angle between gradients is a fundamental quantity in AdvX transferability. Furthermore, we consider the asymmetry of AdvX transferability between two models of the same architecture and explain it in terms of differences in gradient norms between the models. Lastly, we  provide a simple modification to existing training setups that reduces transferability of adversarial examples between pairs of models. 
\end{abstract}

\section{Introduction}

Deep Neural Networks (DNNs) consistently define state-of-the-art performance in many machine learning areas such as image classification, segmentation, speech recognition and language translation \citep{hinton_deep_2012, krizhevsky_imagenet_2012, sutskever_sequence_2014}. These results have lead to DNNs being increasingly deployed in production settings, including self-driving cars, on-the-fly speech translation, and facial recognition for identification. However, like previous machine learning approaches, DNNs have been shown to be vulnerable to adversarial attacks during test time \cite{szegedy_intriguing_2013}. The existence of such AdvX is of immense concern to the security community as human decision making is outsourced to machine learning. Moreover, the arms race of AdvX has been mainly one-sided with attackers being able to fool a diverse variety of proposed defenses.

% \noindent Numerous approaches have been designed to reduce the threat of AdvX. Traditional defense mechanisms are designed with the goal of maximizing the perturbation necessary to trick the network, and making it more obvious to a human eye. However, iterative optimization of adversarial examples by computing gradients in a white box environment or estimating gradients using a surrogate model in a black-box setting have been shown to successfully break such defenses.
% While these methods are theoretically interesting as they can shed light on the nature of potential adversarial attacks, there are many practical applications in which being perceptible to a human is not a reasonable defense against an adversary. For example, in a self-driving car setting, any deep CNN applied to analyzing data originating from a non-visible light spectrum (e.g. LIDAR), could not be protected even by an attentive human observer. 

While numerous approaches have been designed to reduce the threat of AdvX, iterative methods optimizing the AdvX input by computing gradients have to date been successful in breaking proposed defences \citep{carlini_towards_2016}. An unexpected result regarding adversarial examples is their high level of transferability between deep learning models with different architectures and objective functions. In fact, adversarial examples generated for a Deep Convolutional Neural Network (CNN) can even transfer to random forests, SVMs, and logistic regression models \citep{Papernot2016}. This allows attackers to use surrogate models in a black box scenario, estimating gradients to break black box defences. Understanding why these unexpectedly high transferability rates arise could inform future work on both black box and ensemble defences. 

In our work, we focus on the transferability between deep learning models and gradient-based attacks. Specifically, we consider the magnitudes and cosine similarity between input-output Jacobians of model pairs, and show that they are a reliable predictor of transferability. By minimizing the cosine similarity, we are able to significantly reduce the transferability of adversarial examples for any two models we choose. We also explain why the transferability of adversarial examples between two models is asymmetric, and show that it can be made to be symmetric using simple regularization. 

% An unexpected result regarding adversarial examples is their high level of transferability between deep learning models with different architectures and objective functions. In fact, adversarial examples generated for a CNN can even transfer to random forests, SVMs, and logistic regression models. Since the attacks considered in our experiments are gradient-based, we focus on the transferability between deep learning models. Specifically, we consider the cosine similarity between input-output Jacobians of model pairs, and show that it is a reliable predictor of transferability. By minimizing this cosine similarity, we are able to significantly reduce the transferability of adversarial examples for any two models we choose. We also explain why the transferability of adversarial examples between two models is asymmetric, and show that it can be made to be symmetric using simple regularization. 

The contributions of this paper are:%
\begin{itemize}
\item Showing that small input-output Jacobian magnitudes create a false sense of security, and should be reported along with attack parameters%
\item Understanding the asymmetry of adversarial example transferability that still exists when controlling for architecture, hyperparameters, and performance%
\item Introducing a novel regularization scheme for reducing the transferability of adversarial examples between two or more models%
\end{itemize}%
\section{Related Work}
\noindent Adversarial examples in the context of DNNs have come into the spotlight after Szegedy et al. \cite{szegedy_intriguing_2013}, showed the imperceptibility of the perturbations which could fool state-of-the-art computer vision systems. Since then, adversarial examples have been demonstrated in many other domains, notably including speech recognition \cite{carlini_audio_2018}, and  malware detection \cite{grosse_adversarial_2016}. Nevertheless, CNNs in computer vision provide a convenient domain to explore adversarial attacks and defenses, both due to the existence of standardized test datasets, high performing CNN models reaching human or super-human accuracy on clean data, and the marked deterioration of their performance when subjected to adversarial examples to which human vision is robust. 

The intra-model and inter-model transferability of AdvX was investigated thoroughly by \citet{Papernot2016}. It was found that adversarial examples created for fundamentally different MNIST classification models could transfer between each other. This showed that neural networks are not special in their vulnerability, and that simply hiding model details from an attacker is bound to fail since a substitute model can be trained whose AdvX are likely to transfer to the hidden model. Further work has revealed the space of transferable AdvX to be large in dimensionality, and sufficient conditions for transferability were posited as well \cite{tramer_space_2017}. The results support our assumption that defending against transfer-attacks is possible even if the original model is vulnerable to direct attacks. 

Transferable AdvX pose a security risk mainly if they are able to cause a misclassification of the same class between models. Otherwise, a defender could use the disagreement between two models with similar performance as a way of detecting AdvX. \cite{Liu} visualized decision boundaries of several ImageNet classifiers and found that although targeted adversarial examples transfer to other models quite easily, they do not transfer with the same target label. However, a targeted attack on an ensemble of models significantly increases the same-target transfer rate significantly.

It is possible to increase the transferability of AdvX by smoothing the loss function being optimized by an attack as done by \cite{Wu}. The authors use cosine similarity between the gradients of source and target models to justify the claim that increasing the robustness of AdvX to Gaussian perturbations via smoothing increases their transferability. Hence, we also consider cosine similarity between gradients as an important quantity, except that we optimize it directly in order to show that it alone is a significant cause of AdvX transferability. Contrast to Wu et al., we are not optimizing attacks to increase their strength, but rather regularizing models to be more resistant to transfer attacks.  We stress that this is not the same strategy as obfuscating gradients which was shown to give a false sense of security by \cite{athalye_obfuscated_2018}. Instead, we are constraining models to be fundamentally diverse in their robustness to different AdvX. An AdvX that fools one model will not fool the other, and vice versa. This relationship is shown to be attainable empirically via gradient regularization.\vspace{-0.5em}%
\section{Methods}\vspace{-0.5em}
\label{methods}%
\subsection{Attacks}%
We consider three attacks in this paper for completeness. For each attack, we work with the output of the classifier $f$ at the ground truth entry (gt) denoted as $f_{gt}$ instead taking gradients w.r.t. loss functions since it is notationally more convenient.
\subsubsection*{Fast Gradient Sign (FGS)}
We use the following definition of the FGS attack introduced by \citet{goodfellow_explaining_2014}

\begin{align*}
    x' \leftarrow x - \epsilon \mathrm{sign} (\nabla_{x} f_{gt} (x))
\end{align*}

\noindent where $\epsilon$ is a hyperparameter that controls the $\ell_{\infty}$ norm of the perturbation. $\epsilon$ is typically chosen in such as way as to fool the model, but such that the adversarial perturbation is not noticeable to humans, so there is subjectivity involved here as with all attacks. Intuitively, this takes a single step in the direction that minimizes the prediction confidence for the ground truth class of the classifier on input $x$. However, the step is not exactly in the direction of $\nabla_{x} f_{gt}(x)$ since the $\mathrm{sign}$ function can cause a difference in angle between $\nabla_{x} f_{gt}(x)$ and $\mathrm{sign}(\nabla_{x} f_{gt}(x))$ of up to $\frac{\pi}{2}$ in high dimensions. This attack relies on the assumption that neural networks are more or less linear, so just a single step in the gradient direction should be sufficient to cross a decision boundary.

\subsubsection*{Iterative Gradient Sign (IGS)}
The IGS attack \citep{kurakin_adversarial_2016} is very similar to the FGS attack, except that it performs many smaller steps rather than a single step. Two hyperparameters $\epsilon$ and $T$ are used where $\epsilon$ bounds the $\ell_{\infty}$ norm between the final adversarial image and the initial image, and $T$ is the number of iterations. The following procedure is followed%
\begin{align*}
    \alpha &= \frac{\epsilon}{T} \\
    x'_{0} &= x \\
    x'_{t+1} &\leftarrow x'_{t} - \alpha \mathrm{sign} (\nabla_{x} f_{gt} (x))
\end{align*}
\subsubsection*{Carlini and Wagner (CW)}
\citet{carlini_towards_2016} introduced a very powerful attack that solves the following optimization problem using gradient descent:%
\begin{align*}
    \mathrm{minimize} ||\delta||_{p} + c \cdot f(x + \delta)
\end{align*}
Here $c$ is a hyperaparameter that is optimized using binary search and trades off the two simultaneous minimization objectives $||\delta||_{p}$ and $f(x + \delta)$. Plainly, it finds a minimal perturbation that will cause a misclassification.

\subsection{Why Transferability Matters}

AdvX transferability is a significant security concern since it allows adversaries to fool a machine learning service in a black-box scenario. Additionally, transfer attacks can be used in grey-box scenarios where there is a defense in place that obfuscates input-output gradients thereby making it difficult to directly attack the protected model, but simple to build a surrogate model which can be easily attacked and whose adversarial examples transfer to the protected model. Thus, we propose a technique that reduces the transferability between two given models by forcing them to be robust to different sets of AdvX. Given the ability to reduce the transferability between two models, a trivial detection method can be constructed as follows:

\begin{itemize}
    \item Replace the original model $M$ with two models $M_{1}$ and $M_{2}$ such that
        \begin{itemize}
            \item $M_{1}(x) \approx M_{2}(x) \approx M(x), \forall x \in \mathcal{X}_{train}$
            
        \end{itemize}
    \item Compare $|M_{1}(x) - M_{2}(x)|$ to threshold $\kappa$ to classify images as regular or adversarial
\end{itemize}

Intuitively, this is an agreement-based detection method that leverages the fact that $M_{1}$ and $M_{2}$ are robust to different AdvX  in order to flag inputs where the models differ significantly in prediction confidence. In the following sections, we discuss in detail how to reduce AdvX transferability between two models without reducing the performance of those models on the original image classification task. 

\subsection{Jacobian Magnitude Determines Attack Difficulty} \label{sec:jacobian_magnitude}

Basic perturbation-based attacks such as FGS or IGS exploit $\nabla_x f_{gt} (x)$ (input-output gradient) by making changes in input space that are in the opposite direction of $\nabla_x f_{gt} (x)$. More sophisticated attacks such as the CW attack can be intuitively thought of as implicitly using this quantity even though it does not appear in the attack objective itself. We begin by showing a simple result that for a single classifier $f$, $||\nabla_{x} f_{gt} (x)||_{2}$ determines the value of $\epsilon$ required for the FGS attack to fool the classifier. While this result is intuitively obvious and easy to demonstrate, it is often overlooked, and models with a small $||\nabla_{x} f_{gt} (x)||_{2}$ can provide a false sense of security against simple attacks like FGS. 

\noindent The claim that $||\nabla_{x} f_{gt}(x)||_{2}$ controls how large of a perturbation must be made to an input in order to fool a classifier can be demonstrated through a Taylor series approximation.

\noindent Let $a$ be the original input and assume that $f$ outputs probabilities. The most effective single step attack an adversary can perform is to move in the direction opposite of $\nabla_{a} f_{gt}(a)$. Thus, our potential adversarial example is $b = a - \epsilon \nabla_{a} f_{gt} (a)$. The first-order Taylor series approximation of $f_{gt}(x)$ around $a$ is 
\begin{align*}
    f_{gt}(x) &\approx f_{gt}(a) + (x - a) \cdot  \nabla_{a} f_{gt} (a) \\
    f_{gt}(b) &\approx f_{gt}(a) + ((a - \epsilon \nabla_{a} f_{gt}(a)) - a) \cdot \nabla_{a} f_{gt}(a)\\
    &= f_{gt}(a) - \epsilon  \nabla_{a} f_{gt}(a) \cdot \nabla_{a} f_{gt} (a)\\
    &=f_{gt}(a) - \epsilon ||\nabla_{a} f_{gt}(a)||_{2}^{2}
\end{align*}

\noindent Thus, the success of such an attack heavily depends on $||\nabla_{a} f_{gt}(a)||_{2}$. FGS and IGS both apply the signum function to $\nabla_{a} f_{gt}(a)$ as a means of controlling the $\ell_{\infty}$ norm of the adversarial perturbation, but the same principle still applies even in this context. Let $c = a - \epsilon \mathrm{sign} (\nabla_{a} f_{gt} (a))$. Then, 
\begin{align*}
    f_{gt} (c) &\approx f_{gt} (a) + ((a - \epsilon \mathrm{sign} (\nabla_{a} f_{gt} (a))) -a ) \cdot \nabla_{a} f_{gt} (a)\\
    &= f_{gt}(a) - \epsilon \mathrm{sign}(\nabla_{a} f_{gt}(a)) \cdot \nabla_{a} f_{gt} (a)\\
    &= f_{gt}(a) - \epsilon ||\nabla_{a} f_{gt} (a)||_{1}\\
    &\leq f_{gt}(a) - \epsilon ||\nabla_{a} f_{gt} (a)||_{2}
\end{align*}

\noindent Here the last line is due to the triangle inequality. We use this bound instead of regularizing the $\ell_{1}$ norm directly since it seems to be more stable in practice. Given this observation, defense researchers should report $||\nabla_{a} f_{gt}(a)||_{2}$ when they publish new results since reporting a low attack success rate (good defense) can provide a false sense of security against simpler attacks when the gradient magnitude is low.

\begin{figure}[htb]
\centering
\begin{subfigure}{.33\linewidth}
\centering
\includegraphics[width=\linewidth]{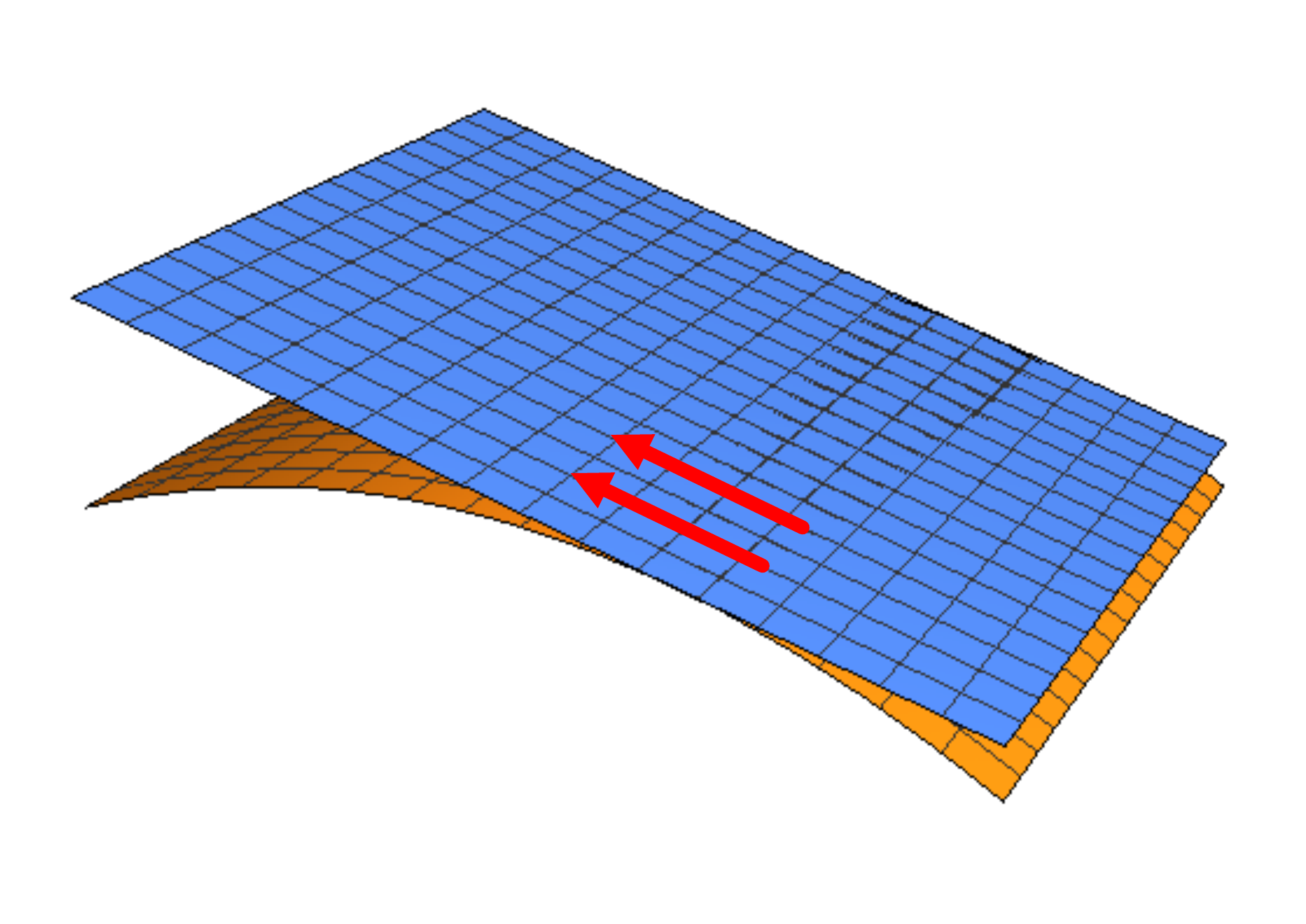}
\caption{Parallel}
\label{fig:sub1}
\end{subfigure}%
\begin{subfigure}{.33\linewidth}
\centering
\includegraphics[width=\linewidth]{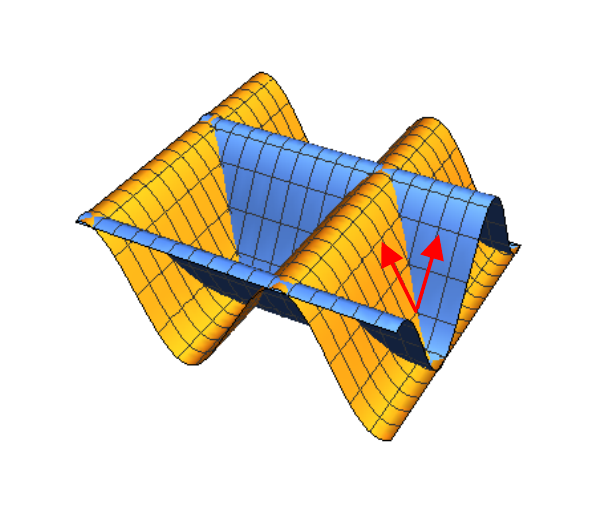}
\caption{Orthogonal}
\label{fig:sub2}
\end{subfigure}
\begin{subfigure}{.33\linewidth}
\centering
\includegraphics[width=\linewidth,height=3cm]{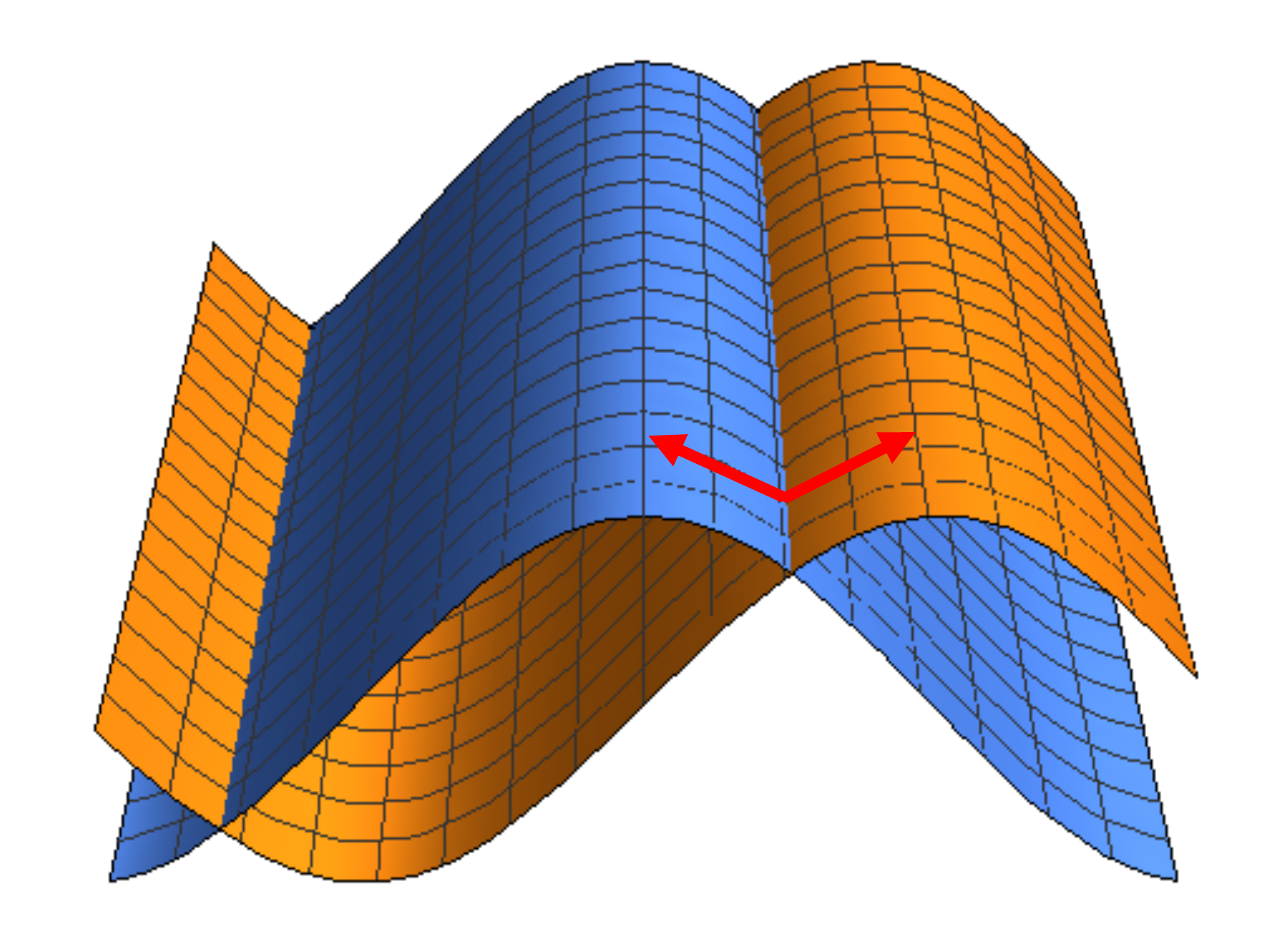}
\caption{Antiparallel}
\label{fig:sub3}
\end{subfigure}
\caption{Illustration of gradient directions for function pairs.}
\label{fig:orthogonal_gradient_direction}
\end{figure}

\subsection{Input-Output Jacobian Magnitude and Transferability Asymmetry}

For two classifiers $f$ and $g$ assume that $f(x)=g(x)$, $\forall x \in \mathcal{X}_{\mathrm{train}}$ where $\mathcal{X}_{\mathrm{train}}$ is some training set. The functions need not be equal when $x \notin \mathcal{X}_{\mathrm{train}}$. If $||\nabla_{x} f_{gt}(x)||_{2} = \alpha ||\nabla_{x} g_{gt}(x)||_{2}$ where $\alpha \in \mathbb{R}, \forall x \in \mathcal{X}_{\mathrm{train}}$, then by the result in the previous section
\begin{align*}
    f_{gt}(c) &\approx f_{gt}(a) - \epsilon ||\nabla_{a} f_{gt} (a)||_{1}\\
    &\leq f_{gt}(a) - \epsilon ||\nabla_{a} f_{gt}(a)||_{2}\\
    &= f_{gt}(a) - \alpha \epsilon ||\nabla_{a} g_{gt}(a)||_{2}
\end{align*}

\noindent Thus, if $\alpha  \not \approx 1$, then $f_{gt}(c) \not \approx g_{gt} (c)$, so we must regularize the models to have $||\nabla_{a} f_{gt}(a)||_{2} \approx ||\nabla_{a} g_{gt}(a)||_{2}$ in order to ensure symmetric transferability results. This may not be a sufficient condition for transfer attack symmetry, but violating this condition is sure to result in asymmetric transferability.
\subsection{Importance of Gradient Angles}

To reduce the transferability of AdvX between two different models, we propose to enforce various geometric relationships between the gradients of the models. Such relationships are visualized in figure \ref{fig:orthogonal_gradient_direction}. For example, for two models $f$ and $g$, if these models happen to be trained such that $\forall x \in \mathcal{X}_{\mathrm{train}}, f_{gt}(x) = g_{gt}(x)$ and $\nabla_{x} f_{gt}(x) \cdot \nabla_{x} g_{gt}(x) = 0$, then the following perturbed input created for $f$, $c = a - \epsilon \nabla_{a} f_{gt}(a)$, will not have an effect on $g$ as shown by the below Taylor series approximation

\begin{align*}
    g_{gt}(c) &\approx g_{gt}(a) + ((a - \epsilon \nabla_{a} f_{gt}(a)) - a) \cdot \nabla_{a} g_{gt}(a)\\
    &= g_{gt}(a) - \epsilon \nabla_{a} f_{gt}(a) \cdot \nabla_{a} g_{gt}(a)\\
    &= g_{gt}(a)
\end{align*}

\noindent This depends on the on the accuracy of the first-order Taylor series approximation around $a$, which can be difficult to determine for neural networks. Unfortunately,  the gradient relationship $\nabla_{x} f_{gt}(x) \cdot \nabla_{x} g_{gt}(x) = 0$ is not sufficient since it implies that the two models have orthogonal gradients which still means that it is possible to simultaneously attack both models by moving in the average direction of the gradients of the two models. More specifically, let $c = a - \epsilon (\frac{\nabla_{a} f_{gt}(a) + \nabla_{a} g_{gt}(a)}{2})$

\begin{align*}
    g_{gt}(c) &\approx g_{gt}(a) + ((a - \frac{\epsilon}{2} (\nabla_{a} f_{gt}(a) + \nabla_{a} g_{gt}(a))) - a) \cdot \nabla_{a} g_{gt}(a)\\
    &= g_{gt}(a) - \frac{\epsilon}{2} (\nabla_{a} f_{gt}(a) \cdot \nabla_{a} g_{gt}(a) + \nabla_{a} g_{gt}(a) \cdot \nabla_{a} g_{gt}(a))\\
    &= g_{gt}(a) - \frac{\epsilon}{2} (||\nabla_{a} g_{gt}(a)||_{2}^{2})
\end{align*}

\noindent So all the orthogonal direction can do against a clever aversary is double the magnitude of the perturbation required to fool the classifier for a single-step attack like FGS. This is likely not sufficient in most cases as the original perturbation is imperceptible to begin with. Thus, we investigate the antiparallel gradient direction induced by the following relationship $\nabla_{x} f_{gt}(x) \cdot \nabla_{x} g_{gt}(x) = -1$. We prove that an adversarial input that uses a convex combination of the gradients cannot simultaneously decrease the prediction confidence of both models. 

Let $r \in [0, 1]$, $c = a - \epsilon (r \nabla_{a} f_{gt}(a) + (1-r) \nabla_{a} g_{gt}(a))$
\begin{align*}
    g_{gt}(c) &\approx g_{gt}(a) + ((a - \epsilon (r \nabla_{a} f_{gt}(a) + (1-r) \nabla_{a} g_{gt}(a))) - a) \cdot \nabla_{a} g_{gt}(a)\\
    &= g_{gt}(a) - \epsilon \underbrace{(r \nabla_{a} f_{gt}(a) \cdot \nabla_{a} g_{gt}(a) + (1 - r) \nabla_{a} g_{gt}(a) \cdot \nabla_{a} g_{gt}(a))}_{\delta}
\end{align*}

\noindent \underline{Case 1: $r < 0.5$}
Assume $||g_{gt}(a)||_{2}^{2} \approx 1$ for convenience which is something that can be attained via regularization but does not remove the generality of this argument. In fact, this is analogous to local Lipschitz-continuity which says that 
\begin{align}
||g(x) - g(y)||_{2} < L \cdot ||x - y||_{2}, \forall x, y  \in B
\end{align}

\noindent where $B$ is simply the union of epsilon balls $C_{i} = \{ y \in R^{N}: ||x_{i} - y|| < \epsilon \}$ for $\epsilon > 0$, and $x_i$ is the i-th element of the training set $\mathcal{X}_{train}$ (note that $B$ is an open set). There exist techniques for making neural networks Lipschitz-continuous \citep{Gouk2018}, but we use simple regularization instead to satisfy the assumption that $||g_{gt}(a)||_{2}^{2} \approx 1$ since the condition is far less restrictive than local Lipschitz-continuity. Thus, 
\begin{align*}
    \delta &= -r + (1-r) ||\nabla_{a} g_{gt}(a)||_{2}^{2}\\
    &\approx 1 - 2r\\
    &> 0
\end{align*}

\noindent This implies that  $g$'s prediction confidence is reduced for the ground truth class. As we can see though, if this input was given to $f$, then 

\begin{align*}
    \delta &= r \nabla_{a} f_{gt}(a) \cdot \nabla_{a} f_{gt}(a) + (1-r) \nabla_{a} g_{gt}(a) \cdot \nabla_{a} f_{gt}(a)\\
    &= r - (1-r)\\
    &= -1 + 2r\\
    &\le 0
\end{align*}

which implies that the prediction confidence of $f$ will actually be increased instead of decreased. We omit the other case of this proof as it is symmetric. \\

Based on the above proof, $\nabla_{x} f_{gt}(x) \cdot \nabla_{x} g_{gt}(x) = -1$ makes it impossible to craft an input via a one-step attack such as FGS that will fool both $f$ and $g$.

\subsection{Obtaining Arbitrary Gradient Relationships} \label{sec:pseudocode}

Any gradient relationship between two models mentioned in the previous section can be obtained by algorithm \ref{alg:training} where parameters are updated in the direction of the negative gradient as to minimize the loss. It is important to note that the outputs of each model should be a vector softmaxed probabilities rather than a vector of logits. For example, if the cosine similarity between between $\nabla_{x} f_{gt}(x)$ and $\nabla_{x} g_{gt}(x)$ is close to 0, but the models output logits, then the output for some other class could still be increasing/decreasing. Alternatively, controlling changes in probabilities does ensure that the probabilities of other classes will not change significantly enough to change the highest probability class since the probabilities have to sum to 1.

\noindent Note that a similar procedure can be used to train a second model once a first model has already been trained instead of training the two models jointly. We acknowledge that this training procedure could result in divergence when optimizing via SGD due to the alternating nature of the updates that switch from optimizing classification loss to gradient loss. Therefore, we also consider a version with a combined loss 
$L_{\mathrm{combined}} = L_{\mathrm{class}} + L_{\mathrm{cos}}$ which updates the model parameters to simultaneously improve both classification and gradient loss. The importance of this consideration is shown in section \ref{sec:simultaneous_updates}.

\begin{algorithm}[t]
  \caption{Proposed Method for Preventing AdvX Transferability 
    \label{alg:training}}  
  \begin{algorithmic} 
    \Require{Assume $M_1$ and $M_2$ are two randomly initialized neural networks}
    \Require{$d(\cdot,\cdot)$ is cosine similarity.}
    \Require{$Q(\cdot,\cdot)$ samples unique minibatches from a dataset}
	\State Let $\mathbf{X, Y}$ denote the training images and labels.
    \State Let $L(M, x, y)$ be cross-entropy of model M
    \State Let $N$ be the number of epochs
    \State $n = 1$
    \While{$n < N$}
    	\For{$Q(\mathbf{x}, \mathbf{y}) \in \mathbf{X, Y}$}
        	\State Compute $L_{\mathrm{class}}=L(M_1, \mathbf{x}, \mathbf{y}) + L(M_2, \mathbf{x}, \mathbf{y})$
            \State Compute $\frac{\partial L_{\mathrm{class}}}{\partial M_1}$ and update $M_1$ params
            \State Compute $\frac{\partial L_{\mathrm{class}}}{\partial M_2}$ and update $M_2$ params
            \State Compute $\mathrm{grad}_1 = \frac{\partial M_1(\mathbf{x})_{\mathrm{\mathbf{y}}}}{x}$
            \Comment $M_1(x)_{\mathrm{y}}$ is a vector of outputs from $M_1$ at true class
            \State Compute $\mathrm{grad}_2 = \frac{\partial M_2(\mathbf{x})_{\mathrm{\mathbf{y}}}}{\mathbf{x}}$
            \State Compute $L_{\mathrm{cos}} = d(\mathrm{grad}_1, \mathrm{grad}_2)$
            \If{goal = perpendicular}
            	\State $L_{\mathrm{cos}} = L_{\mathrm{cos}}^2$
                \Comment Otherwise minimization gives -1 cosine similarity when want 0
            \Else
            	\If {goal = parallel}
            		\State $L_{\mathrm{cos}} = - L_{\mathrm{cos}}$
                    \Comment To change gradient  update to maximize
                \EndIf
            \EndIf
            \State Compute $\frac{\partial L_{cos}}{M_1}$ and update $M_1$ params
            \State Compute $\frac{\partial L_{cos}}{M_2}$ and update $M_2$ params
    	\EndFor
        \State $n = n + 1$
    \EndWhile
    \end{algorithmic}
\end{algorithm}% 
\newpage

\subsection{Verifying Regularization Effectiveness}\label{sec:regularization_effectiveness}
Since it is analytically infeasible to verify where enforced gradient relationships will hold for an arbitrary neural network, we show empirically that the gradient relationships (perpendicular, antiparallel) enforced between two given models on the training set also hold on the test set.

\noindent For tables \ref{table:lenet_mnist_training_grad_stats} and \ref{table:lenet_mnist_test_grad_stats}, 5 pairs of models were trained for all of the relevant gradient similarity scenarios, including one without gradient regularization. The mean column represents the mean cosine similarity across the entire dataset averaged across the 5 model pairs, and the std column represents the standard deviations that were computed for each model pair individually across the entire dataset, and then averaged across all model pairs. It is clear that the gradient regularization works as intended, though achieving perfectly anitparallel gradients is non-trivial. Furthermore, there are some samples for which the cosine similarity between the two models is very far from the goal, but overall the standard deviation is within an acceptable range. Lastly, the L2-norm difference $\left|||\nabla_{x} f_{gt}(x)||_{2} - ||\nabla_{x} g_{gt}(x)||_{2}\right|$ for a given model pair and is an important metric to observe for transferability since it can explain why transferability between two models might be asymmetric. This will be further investigated in section \ref{sec:asymmetry_explained}

\begin{table}[!h]
\caption{Gradient statistics on the training set of LeNet models trained on MNIST.  5 pairs of models were trained for each gradient scenario.}
\label{table:lenet_mnist_training_grad_stats}
\centering
\begin{tabular}{lccc}  
\toprule
Gradient Scenario & Mean Cosine Similarity & Std Cosine Similarity & L2-Norm Difference \\
\midrule
No Regularization & 0.209 $\pm$ 0.017 & 0.001 $\pm$ 0.000 & 0.001 $\pm$ 0.000 \\
Perpendicular      &  0.002 $\pm$ 0.003    & 0.007 $\pm$ 0.003 & 0.007 $\pm$ 0.004    \\
Antiparallel   & -0.912 $\pm$ 0.023    & 0.006 $\pm$ 0.0082 & 0.006 $\pm$ 0.002  \\
\bottomrule
\end{tabular}
\end{table}

\begin{table}[!h]
\caption{Gradient statistics on the test of LeNet models trained on MNIST. 5 pairs of models were trained for each gradient scenario.}
\label{table:lenet_mnist_test_grad_stats}
\centering
\begin{tabular}{lccc}  
\toprule
Gradient Scenario & Mean Cosine Similarity & Std Cosine Similarity & L2-Norm Difference \\
\midrule
No Regularization & 0.209 $\pm$ 0.016 & 0.001 $\pm$ 0.000 & 0.001 $\pm$ 0.000 \\
Perpendicular      & 0.002 $\pm$ 0.003    & 0.007 $\pm$ 0.003 & 0.005 $\pm$ 0.004    \\
Antiparallel   & -0.910 $\pm$ 0.024    & 0.006 $\pm$ 0.002 & 0.006 $\pm$ 002  \\
\bottomrule
\end{tabular}
\end{table}

\noindent As a sanity check, we verify that there is minimal performance loss when using gradient regularization, otherwise the regularization defeats the purpose of the defended system. Table \ref{table:lenet_performance_check} shows the decrease in accuracy for either gradient relation (perpendicular, antiparallel) is within $1\%$ compared to no regularization, so gradient regularization is a reasonable procedure.

\begin{table}[!h]
\caption{Accuracy of LeNet models trained on MNIST. 5 pairs of models were trained for each gradient scenario.}
\label{table:lenet_performance_check}
\centering
\begin{tabular}{lcccc} 
\toprule
\multicolumn{1}{c}{} & \multicolumn{2}{c}{Training Accuracy} &\multicolumn{2}{c}{Test Accuracy} \\
% Gradient Scenario & Training Accuracy & Test Accuracy \\
\cmidrule(r){2-3} \cmidrule(r){4-5}
Gradient Scenario & M1 & M2 & M1 & M2\\
\midrule
No Regularization & 1.000 $\pm$ 0.000 & 1.000 $\pm$ 0.000 & 0.991 $\pm$ 0.001 & 0.992 $\pm$ 0.000\\
Perpendicular      & 1.000 $\pm$ 0.000 & 1.000 $\pm$ 0.000 & 0.989 $\pm$ 0.001 & 0.990 $\pm$ 0.000 \\
Antiparallel   & 0.998 $\pm$ 0.001 & 0.998 $\pm$ 0.002 & 0.981 $\pm$ 0.003 & 0.980 $\pm$ 0.005 \\
\bottomrule
\end{tabular}
\end{table}

\section{Experimental Results}\vspace{-0.5em}

\subsection{Setup}
For section \ref{sec:asymmetry_explained}, 10 models with different random seeds were trained for each gradient magnitude value. Attacks were performed on a random subset of 1000 test images. For section \ref{sec:regularized_transferability_results}, 20 pairs of models were trained for each setting, and attacks were performed on the entire test set. The use of many random seeds was done to obtain confidence intervals and ensure that the observed results are not due to chance. 

All experiments were performed on MNIST. While this could be of concern when demonstrating the effectiveness of a detection method, our principle of gradient regularization has no dependence on the simplicity or complexity of a dataset, nor do we claim that it is a defense for a single model. 

Furthermore, we use the following two-part definition of an adversarial example%
\begin{enumerate}
\item It is classified as a different class than the image from which it was derived.%
\item The image from which it was derived was originally correctly classified.%
\end{enumerate}%
We say that an adversarial example transfers if it satisfies the above definition, but for a different model than the one it was created for.

Additionally, we did not use any data augmentation when training our models since that can be considered as a defense and could further confound results.

All model pairs in sections \ref{sec:regularized_transferability_results} have been regularized such that both models in a pair have a similar input-output Jacobian L2 Norm. This is done in order to allow a univariate analysis of the effect of cosine similarity between model gradients. Otherwise, transferability results could be easily confounded by differences in L2 norms.

Lastly, all attacks performed are untargeted since that is the setting where transferability tends to be highest.

\subsection{Asymmetric AdvX Transferability Explained} \label{sec:asymmetry_explained}
Since there is a clear relationship between $||\nabla_{x} f_{gt} (x)||_{2}$  and attack success rates (section \ref{sec:jacobian_magnitude}), we look at how predictive the difference in magnitudes between the gradients of two models is of transferability rates, i.e. if larger $\left|||\nabla_{x} f_{gt}(x)||_{2} - ||\nabla_{x} g_{gt}(x)||_{2}\right|$ results in more asymmetric transferability. To ensure univariate analysis, we consider models that had the same random initialization, but were trained to have different gradient magnitudes. Cosine similarity is not considered as a factor in this analysis because it is a symmetric metric, and we expect that a low or high cosine similarity would affect transferability between models in a more or less symmetric  way. Table \ref{table:asymmetric} shows how asymmetry in input-output Jacobian magnitude results in asymmetry of AdvX transferability. It is clear that when the models have gradients that are similar in magnitude, AdvX transfer much more symmetrically than when there is a significant difference in magnitude. Transferability decreases quadratically with the grad magnitude of the 2nd model, and the fitted curve has an  $R^2=0.93$ indicating a strong relationship between the two quantities.

\begin{table}[!h]
\caption{Asymmetry in AdvX transferability is at least partly explained by differences in input-output Jacobian magnitude. 10 pairs of LeNet models are considered for each scenario. The models were trained independently, and each model in a pair received the same random weight initialization. The attack being considered is IGS with an epsilon of 1.0}
\label{table:asymmetric}
\centering
\begin{tabular}{lcccc}  
\toprule
Magnitudes & M1 & M2 & M1 to M2 & M2 to M1 \\
\midrule
0.1 \& 0.5     & 0.984 $\pm$ 0.003    & 0.990 $\pm$ 0.001  & 0.824 $\pm$ 0.053 & 0.823 $\pm$ 0.040   \\
% 0.1 \& 0.75     & 0.984 $\pm$ 0.003    & 0.991 $\pm$ 0.001  & 0.833 $\pm$ 0.053 & 0.693 $\pm$ 0.048   \\
0.1 \& 1.0      & 0.984 $\pm$ 0.003    & 0.991 $\pm$ 0.001  & 0.834 $\pm$ 0.055 & 0.560 $\pm$ 0.047   \\
% 0.1 \& 2.0      & 0.984 $\pm$ 0.003    & 0.991 $\pm$ 0.001  & 0.795 $\pm$ 0.053 & 0.381 $\pm$ 0.037   \\
0.1 \& 3.0      & 0.984 $\pm$ 0.003    & 0.990 $\pm$ 0.001  & 0.783 $\pm$ 0.054 & 0.278 $\pm$ 0.034   \\
0.1 \& 5.0      & 0.984 $\pm$ 0.003    & 0.979 $\pm$ 0.010  & 0.757 $\pm$ 0.036 & 0.176 $\pm$ 0.034   \\
0.1 \& 7.0      & 0.984 $\pm$ 0.003    & 0.966 $\pm$ 0.021  & 0.768 $\pm$ 0.041 & 0.143 $\pm$ 0.025   \\
% 2.0 \& 5.0      & 0.991 $\pm$ 0.001    & 0.980 $\pm$ 0.009  & 0.840 $\pm$ 0.044 & 0.447 $\pm$ 0.068   \\
% 2.0 \& 10.0   & 0.967 $\pm$ 0.004    & 0.968 $\pm$ 0.003  & 0.837 $\pm$ 0.090 & 0.829 $\pm$ 0.087    \\
\bottomrule
\end{tabular}
\end{table}

\subsection{Regularized Transferability Results} \label{sec:regularized_transferability_results}

We consider the effects of the various types of gradient regularization mentioned in section \ref{sec:pseudocode} using IGS-1.0 and CW-40. The unconventionally large epsilon value of 1 was selected to obtain large baseline attack success rates such that there is an obvious order of magnitude difference between baseline success rates and transfer rates. The high-confidence CW attack was used instead of the low-confidence version, as per the author's instructions for finding transferable AdvX. Models without gradient regularization were trained to establish what transfer rates are on average in a natural setting. We note that the lowest transferability results for both attacks were achieved by the perpendicular gradients scenario. In fact, the average transfer rate, relative to no regularization, was cut by \textasciitilde56\%  for the IGS attack (table \ref{table:lenet_mnist_igs}), and by \textasciitilde47\% for the CW attack (table \ref{table:lenet_mnist_cw40}) when explicitly training for perpendicular gradients. The high transferability for the antiparallel direction is unexpected, but we consider two possible causes of this. First of all, the models trained were not able to achieve truly antiparallel gradients since the mean cosine similarity was -0.892. Secondly, it is possible that the loss manifold has many critical points, so even if gradients were truly antiparallel locally on training data, moving in the direction that increases confidence for a given model could eventually end up overshooting a peak and ending up in a valley of low confidence. %\todo{make sure these last 3 sentences sound okay}
\begin{table}[htb]
\caption{Transferability results on LeNet model pairs for MNIST attacked with IGS-1.0.}
\label{table:lenet_mnist_igs}
\centering
\begin{tabular}{lcccc}  
\toprule
Gradient Scenario & M1 & M2 & M1 to M2 & M2 to M1 \\
\midrule
No Regularization      & 0.970 $\pm$ 0.038    & 0.981 $\pm$ 0.009  & 0.268 $\pm$ 0.108 & 0.311 $\pm$ 0.125   \\
Parallel All      & 0.863 $\pm$ 0.143    & 0.883 $\pm$ 0.101  & 0.720 $\pm$ 0.180 & 0.741 $\pm$ 0.167   \\
Perpendicular All     & 0.951 $\pm$ 0.077    & 0.946 $\pm$ 0.058  & \bf{0.117 $\pm$ 0.064} & \bf{0.133 $\pm$ 0.046}   \\
Antiparallel All   & 0.869 $\pm$ 0.062    & 0.859 $\pm$ 0.069  & 0.435 $\pm$ 0.110 & 0.487 $\pm$ 0.164    \\
\bottomrule
\end{tabular}
\end{table}
\begin{table}[htb]
\caption{Transferability results on LeNet model pairs for MNIST attacked with CW-40}
\label{table:lenet_mnist_cw40}
\centering
\begin{tabular}{lcccc}  
\toprule
Gradient Scenario & M1 & M2 & M1 to M2 & M2 to M1 \\
\midrule
No Regularization      & 0.991 $\pm$ 0.001    & 0.991 $\pm$ 0.001  & 0.044 $\pm$ 0.028 & 0.048 $\pm$ 0.029   \\
Parallel All     & 0.965 $\pm$ 0.020    & 0.970 $\pm$ 0.014  & 0.696 $\pm$ 0.119 & 0.687 $\pm$ 0.151   \\
Perpendicular All     & 0.987 $\pm$ 0.002    & 0.987 $\pm$ 0.002  & \bf{0.023 $\pm$ 0.011} & \bf{0.022 $\pm$ 0.009}   \\
Antiparallel All  & 0.957 $\pm$ 0.030    & 0.955 $\pm$ 0.031  & 0.456 $\pm$ 0.154 & 0.493 $\pm$ 0.202    \\
\bottomrule
\end{tabular}
\end{table}

\section{Discussion}

Reducing AdvX transferability by using easy-to-implement regularization that is independent of model architecture provides new ways for making model ensembles more robust to AdvX. Attacking an ensemble when the individual models have orthogonal input-output Jacobians is a difficult task since making progress on reducing the confidence of a single model is likely to have little effect on the confidence of another model for the true class. A simple agreement-based detection method for AdvX can be created with just a two-model ensemble which provides an efficient defense that makes no assumptions about data distributions or model architectures. Although the transferability results presented in section \ref{sec:regularized_transferability_results} show a clear improvement in robustness to transfer attacks compared to when no regularization is used, even better results are likely possible. For example, we consider regularizing curvature to further reduce transferability as future work. Note that each model pair had identical architecture and hyperparameters, so it is encouraging that we were able to cut baseline transferability rates by half given this restriction. If different model architectures were used in a pair, in addition to gradient regularization, we believe that AdvX transfer attack rates would be even lower.

\bibliographystyle{unsrtnat}
\bibliography{biblio,Zotero}

\newpage 

\section*{Supplementary Material}

\subsection{Model Details}

\subsubsection*{Individual Models}

Each individual LeNet model trained with regularization that controls input-output Jacobian magnitude had the following architecture:

\begin{verbatim}
LeNet(
  (conv1): Conv2d(1, 6, kernel_size=(3, 3), stride=(1, 1))
  (relu): ReLU()
  (max1): MaxPool2d(kernel_size=2, stride=2, padding=0, dilation=1, 
  					ceil_mode=False)
  (conv2): Conv2d(6, 16, kernel_size=(3, 3), stride=(1, 1))
  (relu): ReLU()
  (max2): MaxPool2d(kernel_size=2, stride=2, padding=0, dilation=1, 
  					ceil_mode=False)
  (fc1): Linear(in_features=400, out_features=120, bias=True)
  (relu): ReLU()
  (fc2): Linear(in_features=120, out_features=84, bias=True)
  (relu): ReLU()
  (fc3): Linear(in_features=84, out_features=10, bias=True)
)
\end{verbatim}

and hyperparameters:

\begin{itemize}
\item Epochs: 500
\item Learning Rate: 0.001
\item Optimizer: Adam with $\beta_{1} = 0.9$ and $\beta_{2} = 0.999$
\item Batch Size: 100
\end{itemize}

Note that there was no shuffling of training data between epochs. There was also no data augmentation, batch normalization, early stopping, etc. This was all done to ensure a univariate analysis when running experiments.

The random seeds used were [100, 109].

\subsubsection*{Model Pairs}

When training LeNet model pairs, the same architecture was used as when training individual models, though the number of epochs was less. The hyperparameters were 

\begin{itemize}
\item Epochs: 200
\item Learning Rate: 0.001
\item Optimizer: Adam with $\beta_{1} = 0.9$ and $\beta_{2} = 0.999$
\item Batch Size: 100
\end{itemize}

Note that since the model weights were randomly initialized sequentially, the two models differed in initialization.

The random seeds used were [150, 169].

\subsection{Importance of Gradient Updates}\label{sec:simultaneous_updates}

To determine the benefit of using simultaneous updates compared to alternating updates, we consider both model performance and gradient cosine similarity at the same time. We do this for the more complex dataset FashionMNIST which allows for a more obvious distinction between the two update techniques. Table \ref{table:lenet_simultaneous_fashion} shows the distinction between the update approaches.  The simultaneous update is more effective since it results in higher accuracy and cosine similarity that is closer to the goal. However, we use the alternating update in our experiments since the benefit is not significant, and because the simultaneous update presents a hyperparameter that needs to trade off classification loss and cosine similarity loss.

\begin{table}[!h]
\caption{Comparison of alternating and simultaneous updates for 5 pairs of LeNet models trained on FashionMNIST. -A indicates alternating, and -S indicates simlutaneous. The improvement in cosine similarity and performance is clear, even though it is small.}
\label{table:lenet_simultaneous_fashion}
\centering
\begin{tabular}{lcccc} 
\toprule
\multicolumn{1}{c}{} & \multicolumn{2}{c}{Accuracy} &\multicolumn{2}{c}{Cosine Similarity} \\
% Gradient Scenario & Training Accuracy & Test Accuracy \\
\cmidrule(r){2-3} \cmidrule(r){4-5}
Gradient Scenario & M1 & M2 & Mean & Std \\
\midrule
Perpendicular -A     & 0.996 $\pm$ 0.002 & 0.994 $\pm$ 0.002 & 0.008 $\pm$ 0.002 & 0.016 $\pm$ 0.001 \\
Perpendicular -S      & 0.997 $\pm$ 0.001 & 0.997 $\pm$ 0.001 & 0.004 $\pm$ 0.001 & 0.013 $\pm$ 0.002 \\
Antiparallel -A  & 0.965 $\pm$ 0.009 & 0.973 $\pm$ 0.007 & -0.908 $\pm$ 0.034 & 0.038 $\pm$ 0.012 \\
Antiparallel -S  & 0.982 $\pm$ 0.013 & 0.979 $\pm$ 0.011 & -0.931 $\pm$ 0.040 & 0.043 $\pm$ 0.015 \\
\bottomrule
\end{tabular}
\end{table}

\end{document}

%% file: math_commands.tex
\usepackage{amsmath,amsfonts,bm}

\def\eqref#1{equation~\ref{#1}}
\def\1{\bm{1}}

\DeclareMathAlphabet{\mathsfit}{\encodingdefault}{\sfdefault}{m}{sl}
\SetMathAlphabet{\mathsfit}{bold}{\encodingdefault}{\sfdefault}{bx}{n}

%% file: main.bbl
\begin{thebibliography}{15}
\providecommand{\natexlab}[1]{#1}
\providecommand{\url}[1]{\texttt{#1}}
\expandafter\ifx\csname urlstyle\endcsname\relax
  \providecommand{\doi}[1]{doi: #1}\else
  \providecommand{\doi}{doi: \begingroup \urlstyle{rm}\Url}\fi

\bibitem[Hinton et~al.(2012)Hinton, Deng, Yu, Dahl, Mohamed, Jaitly, Senior,
  Vanhoucke, Nguyen, Sainath, and Kingsbury]{hinton_deep_2012}
G.~Hinton, L.~Deng, D.~Yu, G.~E. Dahl, A.~r Mohamed, N.~Jaitly, A.~Senior,
  V.~Vanhoucke, P.~Nguyen, T.~N. Sainath, and B.~Kingsbury.
\newblock Deep {Neural} {Networks} for {Acoustic} {Modeling} in {Speech}
  {Recognition}: {The} {Shared} {Views} of {Four} {Research} {Groups}.
\newblock \emph{IEEE Signal Processing Magazine}, 29\penalty0 (6):\penalty0
  82--97, November 2012.
\newblock ISSN 1053-5888.
\newblock \doi{10.1109/MSP.2012.2205597}.

\bibitem[Krizhevsky et~al.(2012)Krizhevsky, Sutskever, and
  Hinton]{krizhevsky_imagenet_2012}
Alex Krizhevsky, Ilya Sutskever, and Geoffrey~E Hinton.
\newblock {ImageNet} {Classification} with {Deep} {Convolutional} {Neural}
  {Networks}.
\newblock In F.~Pereira, C.~J.~C. Burges, L.~Bottou, and K.~Q. Weinberger,
  editors, \emph{Advances in {Neural} {Information} {Processing} {Systems} 25},
  pages 1097--1105. Curran Associates, Inc., 2012.
\newblock URL
  \url{http://papers.nips.cc/paper/4824-imagenet-classification-with-deep-convolutional-neural-networks.pdf}.

\bibitem[Sutskever et~al.(2014)Sutskever, Vinyals, and
  Le]{sutskever_sequence_2014}
Ilya Sutskever, Oriol Vinyals, and Quoc~V Le.
\newblock Sequence to {Sequence} {Learning} with {Neural} {Networks}.
\newblock page~9, 2014.

\bibitem[Szegedy et~al.(2013)Szegedy, Zaremba, Sutskever, Bruna, Erhan,
  Goodfellow, and Fergus]{szegedy_intriguing_2013}
Christian Szegedy, Wojciech Zaremba, Ilya Sutskever, Joan Bruna, Dumitru Erhan,
  Ian Goodfellow, and Rob Fergus.
\newblock Intriguing properties of neural networks.
\newblock \emph{arXiv:1312.6199 [cs]}, December 2013.
\newblock URL \url{http://arxiv.org/abs/1312.6199}.
\newblock arXiv: 1312.6199.

\bibitem[Carlini and Wagner(2016)]{carlini_towards_2016}
Nicholas Carlini and David Wagner.
\newblock Towards {Evaluating} the {Robustness} of {Neural} {Networks}.
\newblock \emph{arXiv:1608.04644 [cs]}, August 2016.
\newblock URL \url{http://arxiv.org/abs/1608.04644}.
\newblock arXiv: 1608.04644.

\bibitem[Papernot et~al.(2016)Papernot, McDaniel, and Goodfellow]{Papernot2016}
Nicolas Papernot, Patrick McDaniel, and Ian Goodfellow.
\newblock {Transferability in Machine Learning: from Phenomena to Black-Box
  Attacks using Adversarial Samples}.
\newblock may 2016.
\newblock URL \url{https://arxiv.org/abs/1605.07277}.

\bibitem[Carlini and Wagner(2018)]{carlini_audio_2018}
Nicholas Carlini and David Wagner.
\newblock Audio {Adversarial} {Examples}: {Targeted} {Attacks} on
  {Speech}-to-{Text}.
\newblock \emph{arXiv:1801.01944 [cs]}, January 2018.
\newblock URL \url{http://arxiv.org/abs/1801.01944}.
\newblock arXiv: 1801.01944.

\bibitem[Grosse et~al.(2016)Grosse, Papernot, Manoharan, Backes, and
  McDaniel]{grosse_adversarial_2016}
Kathrin Grosse, Nicolas Papernot, Praveen Manoharan, Michael Backes, and
  Patrick McDaniel.
\newblock Adversarial {Perturbations} {Against} {Deep} {Neural} {Networks} for
  {Malware} {Classification}.
\newblock \emph{arXiv:1606.04435 [cs]}, June 2016.
\newblock URL \url{http://arxiv.org/abs/1606.04435}.
\newblock arXiv: 1606.04435.

\bibitem[Tramèr et~al.(2017)Tramèr, Papernot, Goodfellow, Boneh, and
  McDaniel]{tramer_space_2017}
Florian Tramèr, Nicolas Papernot, Ian Goodfellow, Dan Boneh, and Patrick
  McDaniel.
\newblock The {Space} of {Transferable} {Adversarial} {Examples}.
\newblock \emph{arXiv:1704.03453 [cs, stat]}, April 2017.
\newblock URL \url{http://arxiv.org/abs/1704.03453}.
\newblock arXiv: 1704.03453.

\bibitem[Liu et~al.(2017)Liu, Chen, Liu, and Song]{Liu}
Yanpei Liu, Xinyun Chen, Chang Liu, and Dawn Song.
\newblock {DELVING INTO TRANSFERABLE ADVERSARIAL EX- AMPLES AND BLACK-BOX
  ATTACKS}.
\newblock 2017.
\newblock URL \url{https://arxiv.org/pdf/1611.02770.pdf}.

\bibitem[Wu et~al.(2018)Wu, Zhu, and Tai]{Wu}
Lei Wu, Zhanxing Zhu, and Cheng Tai.
\newblock {UNDERSTANDING AND ENHANCING THE TRANSFER-ABILITY OF ADVERSARIAL
  EXAMPLES}.
\newblock Technical report, 2018.
\newblock URL \url{https://arxiv.org/pdf/1802.09707.pdf}.

\bibitem[Athalye et~al.(2018)Athalye, Carlini, and
  Wagner]{athalye_obfuscated_2018}
Anish Athalye, Nicholas Carlini, and David Wagner.
\newblock Obfuscated {Gradients} {Give} a {False} {Sense} of {Security}:
  {Circumventing} {Defenses} to {Adversarial} {Examples}.
\newblock \emph{arXiv:1802.00420 [cs]}, February 2018.
\newblock URL \url{http://arxiv.org/abs/1802.00420}.
\newblock arXiv: 1802.00420.

\bibitem[Goodfellow et~al.(2014)Goodfellow, Shlens, and
  Szegedy]{goodfellow_explaining_2014}
Ian~J. Goodfellow, Jonathon Shlens, and Christian Szegedy.
\newblock Explaining and {Harnessing} {Adversarial} {Examples}.
\newblock \emph{arXiv:1412.6572 [cs, stat]}, December 2014.
\newblock URL \url{http://arxiv.org/abs/1412.6572}.
\newblock arXiv: 1412.6572.

\bibitem[Kurakin et~al.(2016)Kurakin, Goodfellow, and
  Bengio]{kurakin_adversarial_2016}
Alexey Kurakin, Ian Goodfellow, and Samy Bengio.
\newblock Adversarial examples in the physical world.
\newblock July 2016.
\newblock URL \url{https://arxiv.org/abs/1607.02533}.

\bibitem[Gouk et~al.(2018)Gouk, Frank, Pfahringer, and Cree]{Gouk2018}
Henry Gouk, Eibe Frank, Bernhard Pfahringer, and Michael Cree.
\newblock {Regularisation of Neural Networks by Enforcing Lipschitz
  Continuity}.
\newblock apr 2018.
\newblock URL \url{http://arxiv.org/abs/1804.04368}.

\end{thebibliography}
